\documentclass[preprint,12pt, a4paper]{elsarticle}

\usepackage{amssymb}
\usepackage{hyperref}

\usepackage{xcolor}
\usepackage{float}
\journal{SoftwareX}

\begin{document}
\renewcommand{\labelenumii}{\arabic{enumi}.\arabic{enumii}}
\newcommand{\glp}{\texttt{gnss\_lib\_py}}
\newcommand{\todo}[1]{{\color{red} #1}}

\begin{frontmatter}
 
\title{gnss\_lib\_py: Analyzing GNSS Data with Python}

\author[label1]{Derek~Knowles}
\author[label2]{Ashwin~Vivek~Kanhere}
\author[label2]{Daniel~Neamati}
\author[label2,label3]{Grace~Gao\corref{cor1}}
\address[label1]{Stanford University, Department of Mechanical Engineering, 440 Escondido Mall
Stanford, CA 94305}
\address[label2]{Stanford University, Department of Aeronautics \& Astronautics, 496 Lomita Mall, Stanford, CA 94305}
\address[label3]{gracegao@stanford.edu}
\cortext[cor1]{corresponding author}

\begin{abstract}

This paper presents \glp{}, a Python library used to parse, analyze, and visualize data from a variety of GNSS (Global Navigation Satellite Systems) data sources. The \glp{} library's ease of use, modular capabilities, testing coverage, and extensive documentation make it an attractive tool not only for scientific and industry users wanting a quick, out-of-the-box solution but also for advanced GNSS users developing new GNSS algorithms. \glp{} has already demonstrated its usefulness and impact through presentation in academic conferences, use in research papers, and adoption in graduate-level university course curricula.

\end{abstract}


\end{frontmatter}

\section*{Metadata}
\label{sec:metadata}

Metadata for the \glp{} library is included in the ancillary data table~\ref{codeMetadata}.

\begin{table}[!h]
\begin{tabular}{|l|p{6.5cm}|p{6.5cm}|}
\hline
C1 & Current code version & v1.0.2 \\
\hline
C2 & Permanent link to code/repository used for this code version & \url{https://github.com/Stanford-NavLab/gnss_lib_py} \\
\hline
C3  & Permanent link to Reproducible Capsule & ---\\
\hline
C4 & Legal Code License   & MIT License\\
\hline
C5 & Code versioning system used & git \\
\hline
C6 & Software code languages, tools, and services used & Python \\
\hline
C7 & Compilation requirements, operating environments \& dependencies & OS: Linux, Mac OS, Windows
\newline Dependencies: numpy, pandas, georinex, unlzw3, pynmea2, pytest, scipy, matplotlib, plotly, requests
\\
\hline
C8 & If available Link to developer documentation/manual &  \url{https://gnss-lib-py.readthedocs.io/} \\
\hline
C9 & Support email for questions & gracegao@stanford.edu \\
\hline
\end{tabular}
\caption{Code metadata}
\label{codeMetadata} 
\end{table}

\section{Motivation and significance}

Global Navigation Satellite Systems (GNSS) are used globally for positioning, navigation, and timing across industries such as transportation, agriculture, power systems, and finance~\cite{morton2021position}.
Several countries and political entities have developed global and regional satellite constellations such as GPS and WAAS (United States), GLONASS (Russia), BeiDou (China), Galileo (the European Union), QZSS (Japan), and IRNSS (India).
GNSS technology, policy, and services are an active research area with established research journals and technical conferences.
The global nature of GNSS means that GNSS research contributions, GNSS measurements, and GNSS datasets also come from sources spread across the globe.
The diverse nature of GNSS resources introduces complexity and confusion when accessing GNSS measurement data, working with multiple datasets, or even comparing algorithms that use GNSS data. For example, different GNSS data sources often use varying conventions for timing or global frames of reference.

Our Python library, \glp{}, lowers the barrier of entry into the GNSS field.
\glp{} is a tool for parsing, analyzing, and visualizing GNSS data for both scientific and industry users, who want a quick, out-of-the-box solution, and GNSS researchers, who create new algorithms and data analysis methods.
\glp{} took inspiration from existing GNSS libraries~\cite{dach2015bernese, rtklib, arizabaleta2021recent, maast, herrera2016gogps, van2017google, laika}.
Previous GNSS libraries were either closed-source; written in MATLAB, C++, or Fortran; or had limited if any documentation on their functionality.
\glp{} is the first GNSS library in Python that is open source, fully documented with interactive tutorials and function-level reference documentation, and completely covered by unit tests.
\glp{} removes the confusion and complexity caused by the different global sources for GNSS data and algorithms by being a single, comprehensive tool that allows users to download GNSS measurements and load GNSS data into a common data structure.
Further, \glp{} contains common baseline algorithms, such as methods for localization and outlier detection, that researchers may use to easily compare against their own novel algorithms. For example, users can import measurements from a GNSS receiver and then, with only a few lines of code, remove outlier measurements and use common methods such as Weighted Least Squares or extended Kalman filtering \cite{morton2021position} to compute a position solution for the receiver.

\glp{} has already contributed towards the process of scientific discovery in research topics such as enabling scalable GNSS outlier detection \cite{knowles2023edmgnss}, predicting whether or not GNSS satellites are line-of-sight when users are in urbanized environments \cite{neamati2023neural}, and mitigating the effects of GPS spoofing \cite{kanhere2023fault}.
Additionally, \glp{} is integrated into the curriculum of two Stanford University graduate-level courses---AA272: Global Positioning Systems and AA275: Navigation for Autonomous Systems.
In these university courses, the teaching staff use \glp{} to create an easy introduction for students to download, analyze, and visualize GNSS data, which enables students to cover a wider array of content and have a more in-depth understanding of the course material.
By using \glp{}, students are able to focus on learning core GNSS concepts rather than spending excess time on tasks such as data retrieval and data processing.
Students also use the supporting functionality of \glp{} in their course projects, where they explore new research directions that align with their interests.

\glp{} has a dedicated documentation website (\href{https://gnss-lib-py.readthedocs.io/}{gnss-lib-py.readthedocs.io}) hosted by readthedocs \cite{readthedocs}. This documentation website has clear instructions for how to install \glp{} on Linux, Windows, MacOS, and in Google Colab.
The documentation website also has detailed tutorials for all \glp{} functionality and function-level docstrings and references for each Python function within the library. In this way, users can easily learn how to work with and debug the code needed to perform their desired task.

\section{Software description}

\begin{figure}[h]
\centering
  \includegraphics[width=\linewidth]{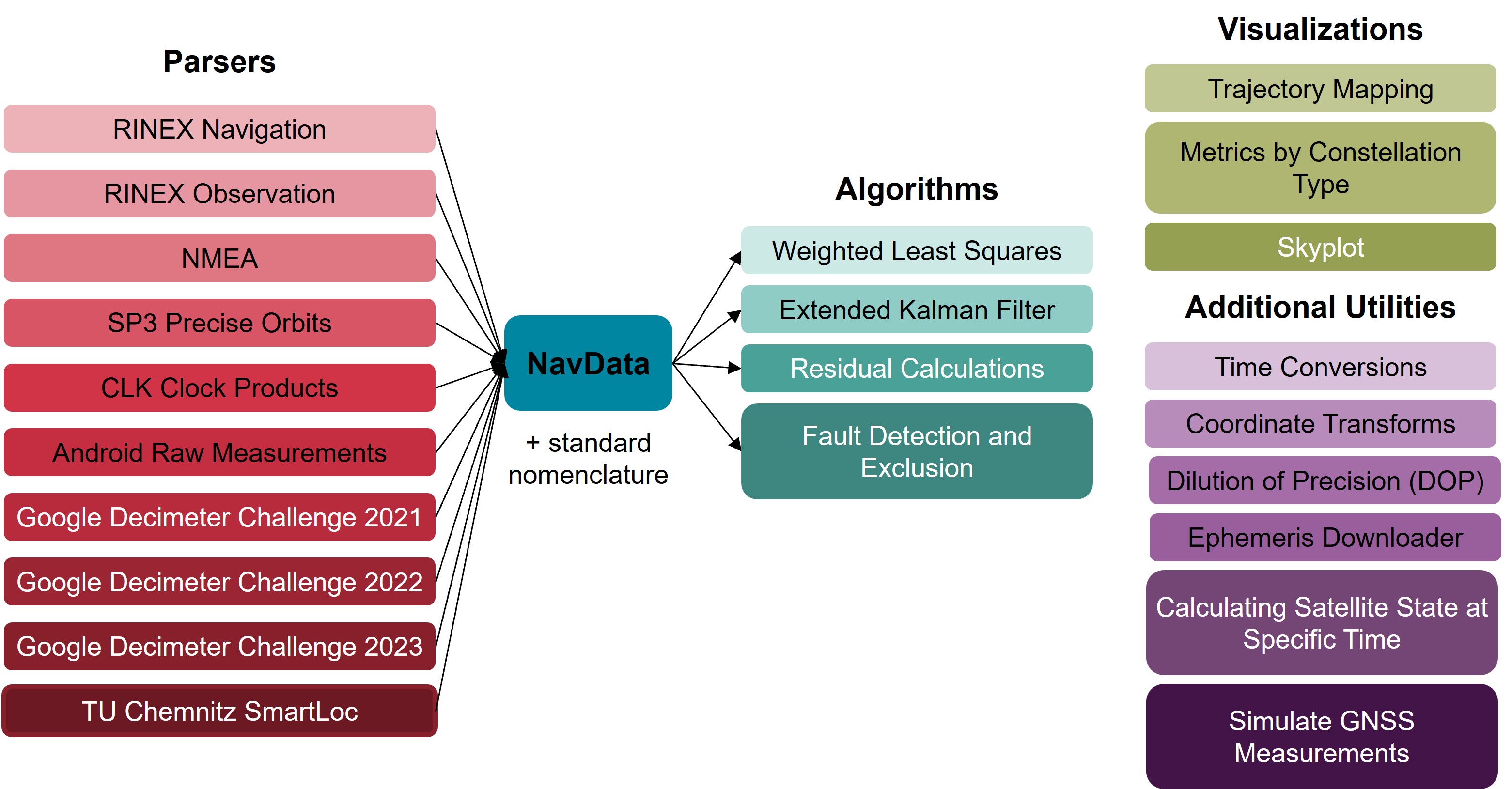}
  \caption{The system architecture of \glp{} is divided into five main categories: parsers, \texttt{NavData}, algorithms, visualizations, and additional utilities.}
  \label{fig:architecture}
\end{figure}

The design principle of \glp{} was to make the library as modular as possible, meaning that many different algorithms can easily be run on a variety of input GNSS data types. 
\glp{} achieves data-type-agnostic modularity by automatically parsing data inputs and transforming them into a common Python class called \texttt{NavData} that uses standard nomenclature.
The \texttt{NavData} Python class allows many different input data types to be used in the same downstream function calls since the data has been set to use common naming conventions, data type, and data structures.
The modular architecture of \glp{} is shown in Figure~\ref{fig:architecture}.
Many different types of measurement data or GNSS datasets can all be parsed and transformed into the common \texttt{NavData} Python class.
Finally, \glp{} implements numerous algorithms, visualization tools, and utility functions that can operate directly on the GNSS data stored in a \texttt{NavData} class object.

\glp{} was also designed to be easily extendable, meaning that new functionality can easily be added in the future to keep the library up-to-date with state-of-the-art GNSS analysis techniques. \glp{} is also fully unit-tested using \texttt{pytest}~\cite{pytest}, ensuring that continued improvements will not break the previously existing functionality of the library.

\subsection{Parsers}
\label{sec:parsers}
The \glp{} library allows users to easily parse through many different types of GNSS measurement data, derived GNSS products, and GNSS datasets. 
\glp{} has prebuilt data parsers for RINEX (Receiver Independent Exchange Format) navigation files that include position and timing data for GNSS satellites as well as RINEX observation files that include raw GNSS measurement observations from reference stations around the globe~\cite{rinex305,rinex400}. 
\glp{} also supports NMEA (National Marine Electronics Association) files, which is a common file type that also contains GNSS measurement data from receivers \cite{nmea_world}. 
\glp{} easily allows users to input derived GNSS products, including orbit parameters using the SP3 (Extended Standard Product - 3) format \cite{sp3}
and satellite clock parameters using the CLK clock format \cite{clk}. 
The final compatible measurement data format that \glp{} currently supports are raw GNSS measurement files from Android phones \cite{fu2020android}. 

Besides GNSS measurement file formats, \glp{} also allows users to input several different common GNSS datasets. \glp{} has pre-built parsers for the 2021, 2022, and 2023 versions of the Google Smartphone Decimeter Challenge \cite{smartphone-decimeter-2021,smartphone-decimeter-2022,smartphone-decimeter-2023}. 
This dataset contains hours of GNSS measurements collected on dozens of different Android phones driving in urban conditions. 
Finally, \glp{} has compatibility with TU Chemnitz's smartLoc GNSS dataset of GNSS receiver data driving in Berlin and Frankfurt \cite{Reisdorf2016}.
To enable future compatibility with new GNSS datasets or data types, \glp{} also has a tutorial for users to make their own custom parser to transfer a new data type into the common \texttt{NavData} Python class using standard naming conventions.

\subsection{\texttt{NavData}}

The \texttt{NavData} Python class offers an intuitive solution for the data structure of GNSS data and is what makes \glp{} modular between many types of data sources and algorithms. The \texttt{NavData} Python class offers the intuition of working with a labeled database while maintaining the speed of vectorized array operations. We have previously shown in \cite{knowles2022modular} that \texttt{NavData}'s implementation with \texttt{NumPy} arrays \cite{numpy} allows even a single matrix operation over data rows to be over two times faster than the same operation on a \texttt{Pandas} dataframe \cite{pandas}.

Users can input data into our custom \texttt{NavData} Python class either as a CSV, \texttt{NumPy} array, \texttt{Pandas} dataframe, or use any of the existing parsers previously discussed in Section~\ref{sec:parsers}. GNSS data sources use a variety of different timing conventions based on UTC time, satellite constellation time, etc., so \glp{} also converts the timing information for all data sources into the number of milliseconds that have elapsed since the start of the GPS epoch on January 6th, 1980. \glp{} calls that time data row ``gps\_millis" and uses ``gps\_millis" as the timing information for all downstream algorithms and utilities. \glp{} also makes adjustments so that satellite and receiver position information is all in an ECEF (Earth-centered, Earth-fixed) reference frame. The standard naming conventions for \glp{} can all be found on the dedicated documentation website \href{https://gnss-lib-py.readthedocs.io/en/latest/reference/reference.html}{https://gnss-lib-py.readthedocs.io/en/latest/reference/reference.html}.

\subsection{Algorithms}
The \glp{} library offers several fundamental algorithms that operate on data which has been parsed into a \texttt{NavData} object.
\glp{} has both temporal filters and snapshot methods for GNSS localization, including an extended Kalman filter and Weighted Least Squares implementation~\cite{morton2021position}.
The \glp{} library also provides algorithms to compute the measurement residuals between the measured and expected pseudorange (the distance between a receiver and a satellite). Finally, \glp{} has fault detection and exclusion algorithms that allow users to remove outlier measurements from GNSS data~\cite{knowles2023edmgnss}.

\subsection{Visualizations}
\glp{} offers several native plotting functions, which allow the user to easily create data visualizations.
Functions are provided to easily access and plot rows of GNSS data, trajectories of GNSS receivers on a map after computing the localization solution, or a skyplot of satellite locations in the sky.

\subsection{Additional Utilities}
To enable the previously mentioned functionality of \glp{}, the library has numerous utility functions listed on the right side of Figure~\ref{fig:architecture}.
\glp{} has functions to convert between different timing conventions and reference frame conventions. 
\glp{} also has the ability to compute the Dilution of Precision of GNSS measurements which is a common metric used to quantify the geometry of the satellite localization problem. 
\glp{} possess a utility tool to automatically download satellite position and timing data called ephemeris for multiple constellations in the form of RINEX navigation files or the derived, precise SP3 and CLK products. 
Currently, members of the GNSS community must manually download the ephemeris files containing satellite information after deciding which source to download from based on the GNSS constellation type(s) and the elapsed time since the measurements were taken. 
Instead, users of the \glp{} library can avoid the hassle of figuring out which data source to download; the user simply inputs the timestamp and constellations of interest into our automatic download tool. 
Finally, \glp{} also has the functionality to use ephemeris data to calculate satellite positions and velocities as well as simulate ionospheric, tropospheric, and other atmospheric effects on the GNSS measurements.

\section{Illustrative example}

\begin{figure}[h]
\centering
  \includegraphics[width=\linewidth]{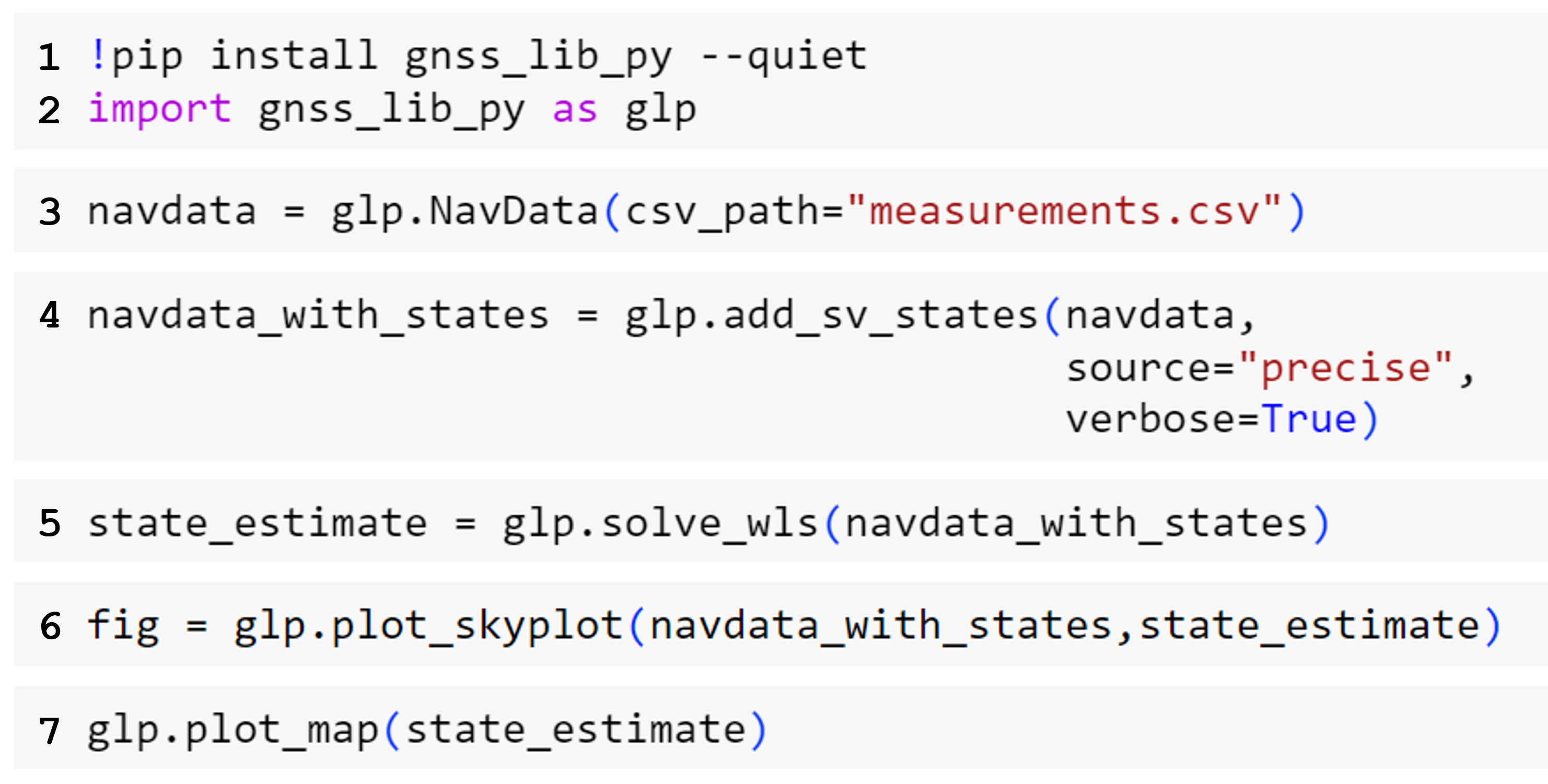}
  \caption{Code needed to calculate a GNSS receiver's position using satellite measurement data.}
  \label{fig:code_snippet}
\end{figure}

In this section, we demonstrate one use case that highlights the ease of using \glp{}. In this example, the user has GNSS measurement data that includes the timestamp, satellite constellation, satellite number, and pseudorange to the satellite. The user wants to use this measurement data to localize their GNSS receiver. Figure~\ref{fig:code_snippet} shows how easy \glp{} makes it to go from GNSS measurement data to a position solution. 

\begin{figure}[ht]

\begin{minipage}[t]{.48\textwidth}
  \centering
  \includegraphics[width=\linewidth]{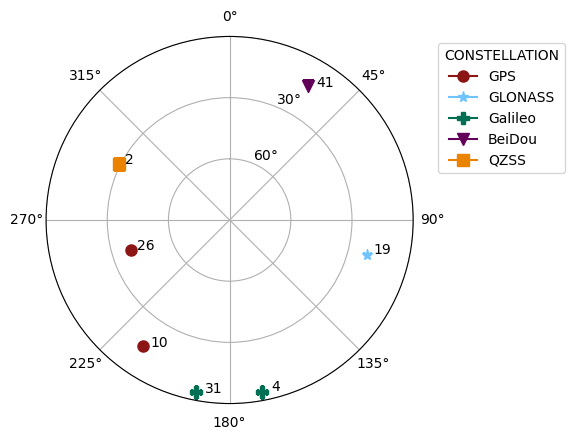}
  \caption{Skyplot of satellite locations.}
  \label{fig:skyplot_timestamp}
\end{minipage}%
\hfill
\begin{minipage}[t]{.42\textwidth}
  \centering
  \includegraphics[width=\linewidth]{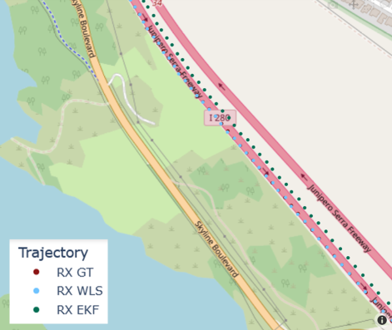}
  \caption{GNSS state estimate data plotted on a map.}
  \label{fig:trajectory}
\end{minipage}
\end{figure}

Lines one and two of the code download the \glp{} library from the Python Package Index using \texttt{pip} and then import the package into the Python workspace.
Line three loads the measurement CSV file into \glp{}'s \texttt{NavData} Python class. To localize the receiver, the user must first know the position and timing information for each satellite from which it received a measurement. Line four adds the satellite position to the \texttt{NavData} class using ``precise" SP3 and CLK satellite data files. Line five then uses the Weighted Least Squares algorithm to solve for the receiver's position. Line six creates the skyplot in Figure~\ref{fig:skyplot_timestamp}, showing the positions in the sky of the satellites from which measurements were received. Finally, line seven of the code can be used to plot the state estimate of the GNSS receiver on an interactive map similar to Figure~\ref{fig:trajectory} that allows the user to qualitatively inspect the position solution of their localization algorithm.

\begin{figure}[h]
\centering
  \includegraphics[width=0.7\linewidth]{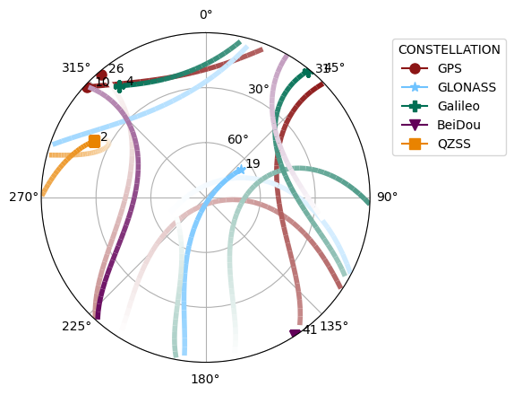}
  \caption{Satellite positions across multiple timesteps.}
  \label{fig:skyplot_full}
\end{figure}

The code snippet from Figure~\ref{fig:code_snippet} demonstrates how \glp{} transforms complicated manual processes into a few simple lines of code.
For example, in order for the function in line four to add the positions of each satellite, it must first download the correct SP3 and CLK satellite ephemeris files using \glp{}'s automatic ephemeris downloader utility.
However, those ephemeris files do not have the satellite positions at the exact timestep for which the user desires.
The ephemeris files have position data for every satellite in the constellation but only in increments of 15 or 30 minutes.
The full satellite position available in an SP3 file can be visualized in Figure~\ref{fig:skyplot_full}. Another \glp{} utility must thus choose the correct satellites to use, then interpolate the satellites' position at the exact millisecond at which the pseudorange measurement was received. The algorithms and utilities are similarly complex for each other line of code shown in Figure~\ref{fig:code_snippet}. The simplicity of this code demonstrates the ease by which even inexperienced GNSS user's can obtain actionable information from GNSS data sources and highlights only one use case out of many for the \glp{} library.

\section{Impact}

The \glp{} library has already been used to process, analyze, and gain insights from GNSS data in multiple research papers. \glp{} has been used in research for detecting and mitigating GPS spoofing \cite{kanhere2023fault}, for detecting and excluding GNSS measurement outliers \cite{knowles2023edmgnss}, and predicting whether or not GNSS satellites are line-of-sight when users are in urbanized environments \cite{neamati2023neural}. \glp{} has also impacted the GNSS community through conference presentations at the International Technical Meeting of the Satellite Division of the Institute of Navigation~\cite{knowles2022modular, knowles2023localization}.
Moreover, \glp{} allows GNSS researchers more time to pursue novel research questions since the researchers no longer have to spend time implementing baseline solutions and instead can quickly compare their algorithms against the baseline solutions implemented by \glp{}.

\glp{} has further impact teaching the next generation of GNSS engineers. \glp{} is used in the curriculum of two Stanford University courses---AA272: Global Positioning Systems and AA 275: Navigation for Autonomous Systems. In these classes, \glp{} is used for both GNSS teaching tutorials as well as by students in class projects researching new topics in GNSS.

Additionally, the \glp{} package has been downloaded hundreds of times a month as demonstrated by statistics from the Python Package Index (\href{https://pypistats.org/packages/gnss-lib-py}{pypistats.org/packages/gnss~-~lib~-~py}).

\section{Conclusions}
\glp{} is modular library used for processing and analyzing GNSS data sources. The ease of use of \glp{} makes it an attractive tool for industry and scientific users who want an out-of-the-box solution for examining GNSS data. At the same time, its modularity and extensive documentation also makes \glp{} a desirable tool for advanced GNSS users who want to extend the functionality of \glp{} or develop new algorithms.

\section*{CRediT authorship contribution statement}

\textbf{Derek Knowles:} Conceptualization, Investigation, Project administration, Software, Validation, Visualization, Writing--original draft, Writing--review \& editing.
\textbf{Ashwin Vivek Kanhere:} Conceptualization, Investigation, Project administration, Software, Validation, Writing--review \& editing.
\textbf{Daniel Neamati:} Investigation, Software, Validation, Writing--review \& editing.
\textbf{Grace Gao:} Supervision, Writing--review \& editing, Resources, Funding acquisition.

\section*{Acknowledgements}
\label{sec:acknowledgements}

The authors would like to thank members of the Stanford NAV Lab who contributed to the \glp{} library, including Dr. Sriramya Bhamidipati, Dr. Shubh Gupta, Adam Dai, Bradley Collicott, Shivam Soni, and Dalton Vega. 

The authors also acknowledge the open-source code repositories on which \glp{} was built and depends including \texttt{NumPy}~\cite{numpy}, \texttt{Pandas}~\cite{pandas}, \texttt{scipy}~\cite{scipy}, \texttt{matplotlib}~\cite{matplotlib}, \texttt{georinex}~\cite{georinex}, \texttt{pytest}~\cite{pytest}, \texttt{unlzw3}, \texttt{pynmea2}, \texttt{plotly}, and \texttt{requests}.

This work was supported by the National Science Foundation (NSF) [award number 2006162] and by NASA Jet Propulsion Laboratory’s (JPL’s) Strategic University Research Partnership program.

\bibliographystyle{elsarticle-num}
\bibliography{references}

\begin{thebibliography}{10}
\expandafter\ifx\csname url\endcsname\relax
  \def\url#1{\texttt{#1}}\fi
\expandafter\ifx\csname urlprefix\endcsname\relax\def\urlprefix{URL }\fi
\expandafter\ifx\csname href\endcsname\relax
  \def\href#1#2{#2} \def\path#1{#1}\fi

\bibitem{morton2021position}
Y.~J. Morton, F.~van Diggelen, J.~J. Spilker~Jr, B.~W. Parkinson, S.~Lo, G.~Gao, Position, navigation, and timing technologies in the 21st century: Integrated satellite navigation, sensor systems, and civil applications, volume 1, John Wiley \& Sons, 2021.

\bibitem{dach2015bernese}
R.~Dach, S.~Lutz, P.~Walser, P.~Fridez, Bernese {GNSS} software version 5.2 (2015).

\bibitem{rtklib}
T.~Takasu, A.~Yasuda, Development of the low-cost {RTK-GPS} receiver with an open source program package {RTKLIB}, in: International symposium on GPS/GNSS, Vol.~1, International Convention Center Jeju Korea, 2009.

\bibitem{arizabaleta2021recent}
M.~Arizabaleta, H.~Ernest, J.~Dampf, T.~Kraus, D.~Sanchez-Morales, D.~D{\"o}tterb{\"o}ck, A.~Sch{\"u}tz, T.~Pany, Recent enhancements of the multi-sensor navigation analysis tool ({M}u{SNAT}), in: Proceedings of the 34th International Technical Meeting of the Satellite Division of The Institute of Navigation (ION GNSS+ 2021), 2021, pp. 2733--2753.
\newblock \href {https://doi.org/10.33012/2021.17960} {\path{doi:10.33012/2021.17960}}.

\bibitem{maast}
T.~Walter, J.~Anderson, A.~Neish, A.~Perkins, \href{https://github.com/stanford-gps-lab/maast}{{MAAST}}.
\newline\urlprefix\url{https://github.com/stanford-gps-lab/maast}

\bibitem{herrera2016gogps}
A.~M. Herrera, H.~F. Suhandri, E.~Realini, M.~Reguzzoni, M.~de~Lacy, go{GPS}: open-source {MATLAB} software, GPS solutions 20~(3) (2016) 595--603.
\newblock \href {https://doi.org/10.1007/s10291-015-0469-x} {\path{doi:10.1007/s10291-015-0469-x}}.

\bibitem{van2017google}
F.~van Diggelen, Google analysis tools for {GNSS} raw measurements, g. co/{GNSST}ools, in: Proceedings of the 30th International Technical Meeting of the Satellite Division of The Institute of Navigation (ION GNSS+ 2017), 2017, pp. 345--356.
\newblock \href {https://doi.org/10.33012/2017.15172} {\path{doi:10.33012/2017.15172}}.

\bibitem{laika}
H.~Schafer, E.~Santana, A.~Haden, R.~Biasini, A commute in data: The comma2k19 dataset, arXiv preprint arXiv:1812.05752 (2018).

\bibitem{knowles2023edmgnss}
D.~Knowles, G.~Gao, \href{https://www.ion.org/publications/abstract.cfm?articleID=19280}{Detection and exclusion of multiple faults using euclidean distance matrices}, in: Proc. of the 36th International Technical Meeting of the Satellite Division of The Institute of Navigation (ION GNSS+ 2023), Denver, CO, 2023, pp. 1774--1782.
\newline\urlprefix\url{https://www.ion.org/publications/abstract.cfm?articleID=19280}

\bibitem{neamati2023neural}
D.~Neamati, S.~Gupta, M.~Partha, G.~Gao, Neural city maps for {GNSS} {NLOS} prediction, in: Proceedings of the 36th International Technical Meeting of the Satellite Division of The Institute of Navigation (ION GNSS+ 2023), 2023, pp. 2073--2087.

\bibitem{kanhere2023fault}
A.~V. Kanhere, G.~Gao, Fault-robust {GPS} spoofing mitigation with expectation-maximization, in: Proceedings of the 36th International Technical Meeting of the Satellite Division of The Institute of Navigation (ION GNSS+ 2023), 2023, pp. 3815--3828.

\bibitem{readthedocs}
E.~Holscher, C.~Leifer, B.~Grace, Read the docs, URL< https://docs. readthedocs. org (2010).

\bibitem{pytest}
H.~Krekel, B.~Oliveira, R.~Pfannschmidt, F.~Bruynooghe, B.~Laugher, F.~Bruhin, \href{https://github.com/pytest-dev/pytest}{pytest 6.2.5} (2004).
\newline\urlprefix\url{https://github.com/pytest-dev/pytest}

\bibitem{rinex305}
I.~Romero, \href{https://files.igs.org/pub/data/format/rinex305.pdf}{The receiver independent exchange format version 3.05} (2020).
\newline\urlprefix\url{https://files.igs.org/pub/data/format/rinex305.pdf}

\bibitem{rinex400}
I.~Romero, \href{https://files.igs.org/pub/data/format/rinex_4.00.pdf}{The receiver independent exchange format version 4.00} (2021).
\newline\urlprefix\url{https://files.igs.org/pub/data/format/rinex_4.00.pdf}

\bibitem{nmea_world}
E.~Gakstatter, \href{https://www.gpsworld.com/what-exactly-is-gps-nmea-data/}{What exactly is {GPS} {NMEA} data?}, {GPS} World (2015).
\newline\urlprefix\url{https://www.gpsworld.com/what-exactly-is-gps-nmea-data/}

\bibitem{sp3}
P.~Spofford, B.~Remondi, \href{https://files.igs.org/pub/data/format/sp3_docu.txt}{The national geodetic survey standard gps format {SP3}} (1994).
\newline\urlprefix\url{https://files.igs.org/pub/data/format/sp3_docu.txt}

\bibitem{clk}
J.~Ray, W.~Gurtner, \href{https://files.igs.org/pub/data/format/rinex_clock300.txt}{{RINEX} extensions to handle clock information v3.00} (2006).
\newline\urlprefix\url{https://files.igs.org/pub/data/format/rinex_clock300.txt}

\bibitem{fu2020android}
G.~M. Fu, M.~Khider, F.~van Diggelen, Android raw {GNSS} measurement datasets for precise positioning, in: Proceedings of the 33rd international technical meeting of the satellite division of the Institute of Navigation (ION GNSS+ 2020), 2020, pp. 1925--1937.
\newblock \href {https://doi.org/10.33012/2020.17628} {\path{doi:10.33012/2020.17628}}.

\bibitem{smartphone-decimeter-2021}
D.~Orendorff, F.~van Diggelen, J.~Elliott, M.~Fu, M.~Khider, S.~Dane, \href{https://kaggle.com/competitions/google-smartphone-decimeter-challenge}{Google smartphone decimeter challenge} (2021).
\newline\urlprefix\url{https://kaggle.com/competitions/google-smartphone-decimeter-challenge}

\bibitem{smartphone-decimeter-2022}
A.~Howard, A.~Chow, B.~Julian, D.~Orendorff, M.~Fu, K.~Mohammed, S.~Dane, \href{https://kaggle.com/competitions/smartphone-decimeter-2022}{Google smartphone decimeter challenge 2022} (2022).
\newline\urlprefix\url{https://kaggle.com/competitions/smartphone-decimeter-2022}

\bibitem{smartphone-decimeter-2023}
A.~Chow, D.~Orendorff, M.~Fu, K.~Mohammed, S.~Dane, V.~Gulati, \href{https://kaggle.com/competitions/smartphone-decimeter-2023}{Google smartphone decimeter challenge 2023} (2023).
\newline\urlprefix\url{https://kaggle.com/competitions/smartphone-decimeter-2023}

\bibitem{Reisdorf2016}
P.~Reisdorf, T.~Pfeifer, J.~Bre{\ss}ler, S.~Bauer, P.~Weissig, S.~Lange, G.~Wanielik, P.~Protzel, The problem of comparable {GNSS} results--an approach for a uniform dataset with low-cost and reference data, in: M.~Ullmann, K.~El-Khatib (Eds.), The Fifth International Conference on Advances in Vehicular Systems, Technologies and Applications, Vol.~5, 2016, p.~8, iSSN: 2327-2058.

\bibitem{knowles2022modular}
D.~Knowles, A.~V. Kanhere, S.~Bhamidipati, G.~Gao, A modular and extendable gnss python library, in: Proceedings of the 35th International Technical Meeting of the Satellite Division of The Institute of Navigation (ION GNSS+ 2022), 2022, pp. 883--888.

\bibitem{numpy}
C.~R. Harris, K.~J. Millman, S.~J. van~der Walt, R.~Gommers, P.~Virtanen, D.~Cournapeau, E.~Wieser, J.~Taylor, S.~Berg, N.~J. Smith, R.~Kern, M.~Picus, S.~Hoyer, M.~H. van Kerkwijk, M.~Brett, A.~Haldane, J.~Fernández~del Río, M.~Wiebe, P.~Peterson, P.~Gérard-Marchant, K.~Sheppard, T.~Reddy, W.~Weckesser, H.~Abbasi, C.~Gohlke, T.~E. Oliphant, Array programming with {NumPy}, Nature 585 (2020) 357–362.
\newblock \href {https://doi.org/10.1038/s41586-020-2649-2} {\path{doi:10.1038/s41586-020-2649-2}}.

\bibitem{pandas}
{The pandas development team}, \href{https://github.com/pandas-dev/pandas}{{pandas-dev/pandas: Pandas}}.
\newline\urlprefix\url{https://github.com/pandas-dev/pandas}

\bibitem{knowles2023localization}
D.~Knowles, A.~V. Kanhere, G.~Gao, Localization and fault detection baselines from an open-source python gnss library, in: Proceedings of the 36th International Technical Meeting of the Satellite Division of The Institute of Navigation (ION GNSS+ 2023), 2023, pp. 1565--1571.

\bibitem{scipy}
P.~Virtanen, R.~Gommers, T.~E. Oliphant, M.~Haberland, T.~Reddy, D.~Cournapeau, E.~Burovski, P.~Peterson, W.~Weckesser, J.~Bright, S.~J. {van der Walt}, M.~Brett, J.~Wilson, K.~J. Millman, N.~Mayorov, A.~R.~J. Nelson, E.~Jones, R.~Kern, E.~Larson, C.~J. Carey, {\.I}.~Polat, Y.~Feng, E.~W. Moore, J.~{VanderPlas}, D.~Laxalde, J.~Perktold, R.~Cimrman, I.~Henriksen, E.~A. Quintero, C.~R. Harris, A.~M. Archibald, A.~H. Ribeiro, F.~Pedregosa, P.~{van Mulbregt}, {SciPy 1.0 Contributors}, {{SciPy} 1.0: Fundamental Algorithms for Scientific Computing in Python}, Nature Methods 17 (2020) 261--272.
\newblock \href {https://doi.org/10.1038/s41592-019-0686-2} {\path{doi:10.1038/s41592-019-0686-2}}.

\bibitem{matplotlib}
J.~D. Hunter, Matplotlib: A 2{D} graphics environment, Computing in Science \& Engineering 9~(3) (2007) 90--95.
\newblock \href {https://doi.org/10.1109/MCSE.2007.55} {\path{doi:10.1109/MCSE.2007.55}}.

\bibitem{georinex}
\href{https://doi.org/10.5281/zenodo.3401498}{{georinex}} (2022).
\newline\urlprefix\url{https://doi.org/10.5281/zenodo.3401498}

\end{thebibliography}

\end{document}